\documentclass[runningheads]{llncs}

\usepackage{graphicx}
\usepackage{booktabs}
\usepackage{amsmath,amssymb,amsfonts}
\usepackage{algorithmic}
\usepackage{graphicx}
\usepackage{textcomp}
\usepackage{enumerate}
\usepackage{booktabs}
\usepackage{calc}
\usepackage{subfigure}
\usepackage{wrapfig}
\usepackage[keeplastbox]{flushend}
\usepackage[table]{xcolor}
\usepackage{colortbl}
\usepackage{framed}
\usepackage[framemethod=TikZ]{mdframed}
\usepackage{multirow}
\usepackage{mathrsfs}
\usepackage[ruled,vlined]{algorithm2e}
\usepackage{appendix}
\usepackage{balance}
\usepackage{bm}
\usepackage{diagbox}
\usepackage[misc]{ifsym}

\newlength\colomnWidth
\begin{document}

\title{
An Approach for Process Model Extraction By Multi-Grained Text Classification
}

\author{Chen Qian$^1$ \and Lijie Wen$^{1(\text{\Letter})}$ \and Akhil Kumar$^2$ \and Leilei Lin$^1$ \and Li Lin$^1$ \and Zan Zong$^1$ \and Shu$'$ang Li$^1$ \and Jianmin Wang$^1$}

\institute{
School of Software, Tsinghua University, Beijing 100084, China \and
Smeal College of Business, Penn State University, 16802 State College, USA\\
\email{qc16@mails.tsinghua.edu.cn, wenlj@tsinghua.edu.cn}
}

\maketitle

\begin{abstract}
Process model extraction (PME) is a recently emerged interdiscipline between natural language processing (NLP) and business process management (BPM), which aims to extract process models from textual descriptions. Previous process extractors heavily depend on manual features and ignore the potential relations between clues of different text granularities. In this paper, we formalize the PME task into the multi-grained text classification problem, and propose a hierarchical neural network to effectively model and extract multi-grained information without manually-defined procedural features. Under this structure, we accordingly propose the coarse-to-fine (grained) learning mechanism, training multi-grained tasks in coarse-to-fine grained order to share the high-level knowledge for the low-level tasks. To evaluate our approach, we construct two multi-grained datasets from two different domains and conduct extensive experiments from different dimensions. The experimental results demonstrate that our approach outperforms the state-of-the-art methods with statistical significance and further investigations demonstrate its effectiveness.

\end{abstract}

\section{Introduction} \label{sec:introduction}
\begin{figure}[t]
\centering
\includegraphics[width=0.99\columnwidth]{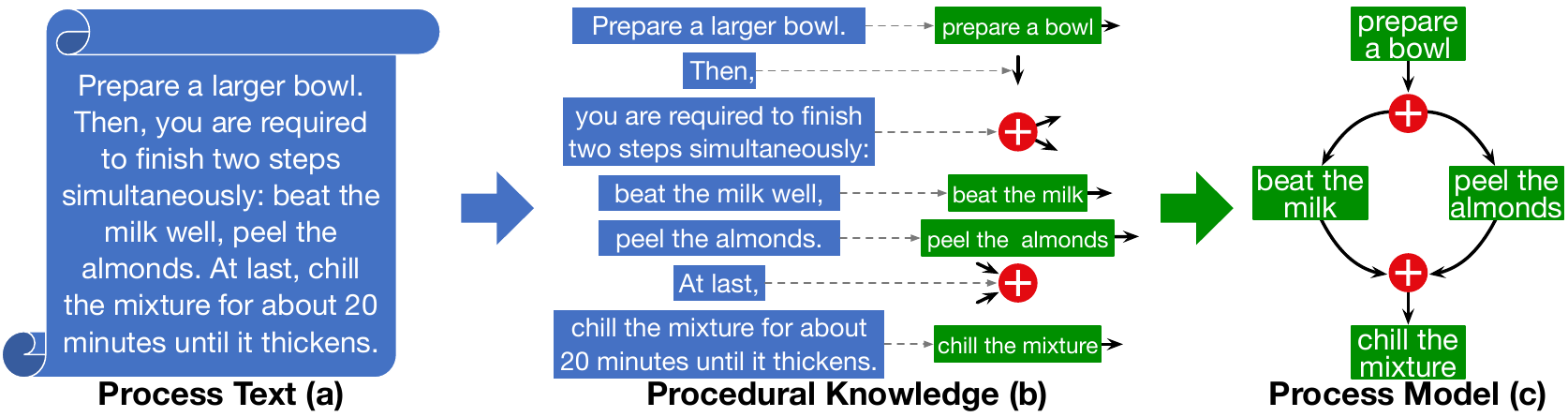}
\caption{Illustration of the PME problem.}
\label{fig:example}
\end{figure}

The widespread adoption of conversational agents such as Alexa, Siri and Google Home demonstrates the natural demand for such assistive agents. To go beyond supporting the simplistic queries such as ``\textit{what should I do next?}'', these agents need domain-specific procedural knowledge \cite{extracting_action_sequences}. Procedural knowledge, also called ``\emph{how-to-do-it}'' knowledge, is the knowledge related to the execution of a series of interrelated tasks \cite{extracting_control_flow}. A major source of procedural knowledge is contained in natural textual instructions \cite{automatically_extracting}, such as cooking recipes that describe cooking procedures and maintenance manuals that describe repair procedures for various devices and gadgets. While it is possible to manually understand, extract and reuse such knowledge from texts, ultimately that is a very labor-intensive option \cite{identifying_candidate_tasks}. To facilitate reuse and repurpose of procedural knowledge, \textbf{process model extraction (PME)} is emerging to automatically extract underlying \textbf{process models} from \textbf{process texts}. As Figure \ref{fig:example} illustrates, PME extracts and presents the main actions (nodes) and their ordering relations (sequence flows) expressed in the cooking recipe as a process model. This task can liberate humans from the manual efforts of creating and visualizing procedural knowledge by making assistive agents understand process texts intelligently \cite{extracting_action_sequences}.

However, PME is challenging as it requires agents to understand complex descriptions of actions and involves multi-grained information mining. For example, in Figure \ref{fig:example}, to extract the whole process model, the process extractor has to recognize \textit{whether a sentence is describing an action}? (sentence-level information) and \textit{who does what in a sentence describing an action}? (word-level information). Recent research efforts have been made to extract the main procedural knowledge. For example, language-rule based extractors \cite{ontology_based_workflow_extraction} used pre-defined language rules to extract procedural knowledge. Pattern-matching based extractors \cite{extracting_control_flow} used NLP tagging or parsing tools to analyze sentences and extract corresponding information from triggered patterns. Knowledge-based extractors employed pre-defined ontology \cite{ontology_based_workflow_extraction} or world model \cite{process_model_generation} to help extract that information. However, traditional methods suffer from weak generalizability when applied in open-domain or open-topic scenarios since they: 1) require much large-scale and domain-specific procedural features; and 2) ignore the relations between sentence-level and word-level subtasks.

In this paper, we propose a new framework to extract process models from process texts. Specifically, we first formalize the PME task into the multi-grained text classification problem, and then propose a hierarchical neural network to effectively model and extract multi-grained information without manually defined procedural features. Under this hierarchical structure, we accordingly propose the coarse-to-fine (grained) learning mechanism, training multi-grained tasks in coarse-to-fine grained order, to share the high-level knowledge for the low-level tasks. To train and evaluate our model, we construct two multi-grained datasets from two different domains and conduct extensive experiments from different dimensions. Experimental results demonstrate that our approach outperforms state-of-the-art methods with statistical significance.

In summary, this paper makes the following contributions:
\begin{enumerate}[$\bullet$]
\item We first formalize the PME task into the multi-grained text classification problem and design a new hierarchical network to model the conditional relation among multi-grained tasks. Supported by automatic feature extraction, it can extract procedural knowledge without employing manual features and defining procedural knowledge.
\item We propose the coarse-to-fine learning mechanism that trains multi-grained tasks in coarse-to-fine (grained) order to apply the sentence-level knowledge for the word-level tasks.
\item We construct two multi-grained datasets from two different domains to train and evaluate multi-grained text classifiers. The results demonstrate that our approach outperforms the state-of-the-art methods.
\end{enumerate}

\section{Related Work}
Several \textbf{language-rule based} methods have been originally applied to process extraction \cite{automatically_extracting,workflow_extraction_from,extraction_of_procedural_knowledge,creating_and_updating}. Specifically, \cite{automatically_extracting} introduced a generic semantic representation of procedures for analyzing instructions, using Stanford Parser to automatically extract structured procedures from instructions. \cite{workflow_extraction_from} described an approach for the automatic extraction of workflows from cooking recipes resulting in a formal description of cooking instructions. A chain of standard information extraction pipeline was applied with the help of GATE. They were dedicated to the special characteristics of textual cooking instructions (verb centric, restricted vocabulary of ingredients, relatively independent sentences). Although they are easy to develop and interpret, they require a large number of linguistic rules created by domain experts.

Along this line, \textbf{pattern-matching based} methods \cite{extraction_of_process_models,extracting_and_enriching,on_the_Use,extracting_control_flow,automatic_process_model,process_model_generation} designed various language patterns, considering basic language patterns \cite{extracting_and_enriching}, syntactic tree \cite{extraction_of_process_models,automatic_process_model} and anaphora resolution \cite{extracting_control_flow,on_the_Use}. For example, \cite{extracting_and_enriching} presented on the step of anaphora resolution to enrich the process models extracted by introducing a lexical approach and two further approaches based on a set of association rules which were created during a statistical analysis of a corpus of workflows. However, these studies are to some extent limited by domain-specific knowledge bases, making them not applicable for open-domain or open-topic scenarios.

Recently, some \textbf{knowledge based} methods \cite{automated_generation,ontology_based_workflow_extraction,bioinformatic_workflow_extraction} have been applied to this problem and they were shown to perform well. For example, \cite{ontology_based_workflow_extraction} proposed an ontology-based workflow extraction framework that extended classic NLP techniques to extract and disambiguate tasks in texts. Using a model-based representation of workflows and a domain ontology, the extraction process used a context-based approach to recognize workflow components such as data and control elements in a flow. However, they also require a large quantity of cognition-level knowledge, such as a world model \cite{automated_generation,process_model_generation} or an ontology \cite{ontology_based_workflow_extraction,bioinformatic_workflow_extraction}, which would be time-consuming and labor-intensive to build.

There also exist some \textbf{machine-learning based} studies which incorporated traditional machine learning techniques into process extraction \cite{identifying_candidate_tasks,case_information_extraction}. \cite{identifying_candidate_tasks} leveraged support vector machine to automatically identify whether a task described in a textual process description is manual or automated. \cite{case_information_extraction} used semi-supervised conditional random fields and support vector machine to label process texts and recognize main information. 

Other process-related works include process state tracing, action extraction and process search. For example, \cite{Building_Dynamic_Knowledge,Tracking_State_Changes,Reasoning_about_Actions} proposed neural machine-reading models that constructed dynamic knowledge graphs from procedural text. \cite{Simulating_Action_Dynamics} introduced a network to understand the procedural text through simulation of action dynamics. \cite{BePT,Goun} set out to generate the textual descriptions from process models. \cite{Mise_en_Place} proposed a probabilistic model to incorporate aspects of procedural semantics and world knowledge. \cite{Modeling_Biological_Processes} aimed to answer biological questions by predicting a rich process structure and mapping the question to a formal query. 


\section{Methodology}

We first formalize the PME task into the multi-grained text classification problem. Given a process text $\mathcal{T}=\langle S^1, S^2, \cdots, S^n \rangle$ where $S^i=\langle W^i_1, W^i_2, \cdots, W^i_{\vert S^i \vert} \rangle$ is a sequence of words, $\forall i=1,2,\cdots,n$. For each $S^i \in \mathcal{T}$, the key of PME ($\alpha$) is to predict a corresponding set of labels that describe the type of the sentence and its corresponding word-level arguments, $\alpha(S^i)$=($sType$, $sSemantic$, $sArgs$). $sType$ indicates the type of sentence $S^i$, which can be $Action$ or $Statement$. The $Action$ indicates that $S^i$ is an action mention and $Statement$ refers to a non-action sentence. If $S^i$ is categorized into $Action$ type, then $sSemantic$ is $\varnothing$ (empty marker) and $sArgs$=[$aRole$, $aName$, $aObject$] denotes the action's executor, action name and direct object respectively. Otherwise, $sArgs$ is $\varnothing$ and $sSemantic$ can be one of $\{ \triangleright, \triangleleft, \bullet, \times, + \}$ relation symbols that determine how actions are coordinated. The five relations refer to the beginning of a block of actions (\textit{block begins}), the ending of a block of actions (\textit{block ends}), a successive relation, an optional relation and a concurrent relation, respectively.

\begin{example}\label{exm:example}
Consider the two sentences in Figure \ref{fig:example}: $S^i$=``you are required to finish two steps'' and $S^j$=``chill the mixture for about 20 minutes until it thickens''. $S^i$ means that two following actions should be chosen and done concurrently, thus it is labeled as a \textit{concurrency relation}, $\alpha(S^i)=(Statement, +, \varnothing)$. $S^j$ is an \textit{action mention}, thus it is labeled with its role, action name and object, $\alpha(S^j)=(Action, \varnothing, [\varnothing, chill, mixture])$.
\end{example}

We argue that PME involves three main text classification subtasks: 
\begin{enumerate}[{\bf ST1}]
\item \textbf{Sentence Classification} (Sentence-level): identifying whether a sentence is describing an action or a statement.
\item \textbf{Sentence Semantics Recognition} (Sentence-level): recognizing the semantics of a $Statement$ sentence to control the execution of following actions, i.e., \textit{block begins}, \textit{block ends}, \textit{successive relation}, \textit{optional relation} and \textit{concurrency relation}.
\item \textbf{Semantic Role Labeling} (Word-level): assigning semantic roles ($aRole$, $aName$, $aObject$) to words or phrases in an $Action$ sentence.
\end{enumerate}
Note that these three tasks are not independent because ST2 and ST3 are conditioned on ST1, i.e., for a single sentence, the result of ST1 determines whether the sentence is passed to ST2 or ST3 .

\subsection{Overall Framework}

\begin{figure*}[t]
\centering
\includegraphics[width=0.99\columnwidth]{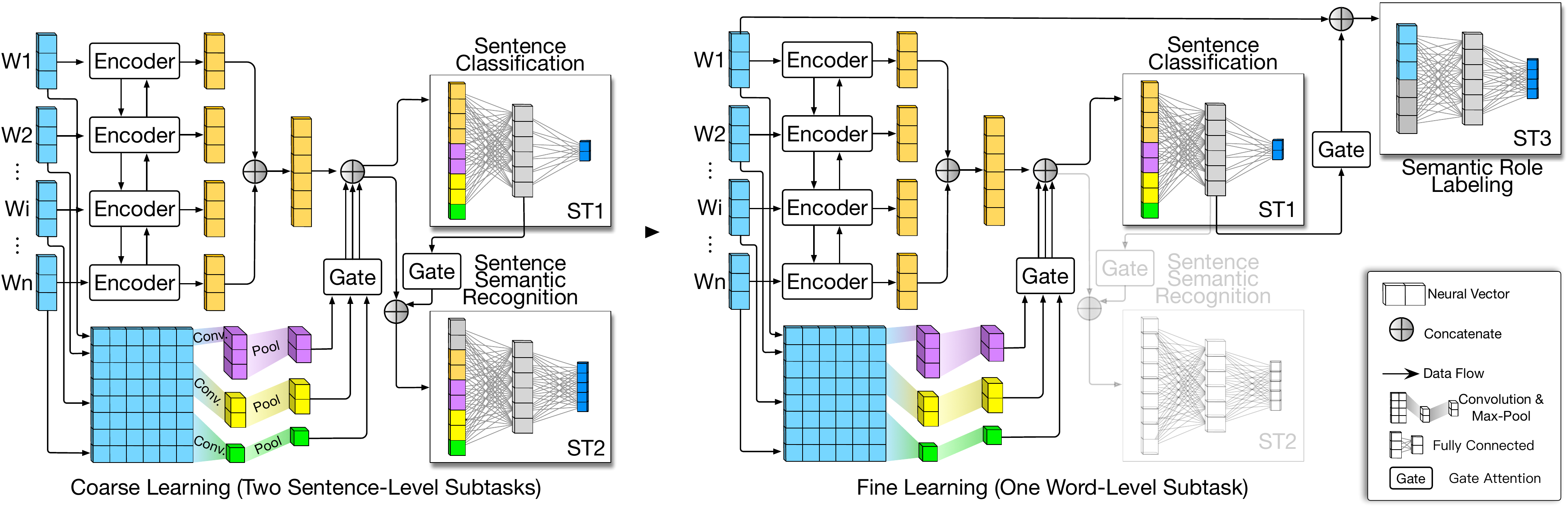}
\caption{High-level overview of the proposed method MGTC, which consists of two stages: a) The coarse-grained learning phase combines the bi-directional encoding layer and convolutional neural networks to \textit{pre-train} two sentence-level tasks. b) The fine-grained learning phase utilizes the learned sentence-level knowledge to \textit{fine-tune} the word-level task.}
\label{fig:Framework}
\end{figure*}

To mine sentence-level and word-level information effectively, we propose a deep-learning-based network to effectively avoid manually defining domain-specific procedural features, called \underline{M}ulti-\underline{G}rained \underline{T}ext \underline{C}lassifier (MGTC). Figure \ref{fig:Framework} shows the framework of our proposed approach. Specifically, we design a hierarchical structure to model the conditional relations between three subtasks and to effectively extract textual clues from different granularities. Under this framework, we accordingly propose the coarse-to-fine (grained) learning mechanism, training coarse tasks in advance before training fine-grained tasks to share the learned high-level knowledge to the low-level tasks, which conforms with the procedure of human learning more than those methods without the consideration of different granularities. By problem mapping, the output of our model is the sentence-level and word-level labels (see Figure \ref{fig:example}(b) and Example \ref{exm:example}), which could be further easily visualized (i.e., from Figure \ref{fig:example}(b) to Figure \ref{fig:example}(c)) as a specific/general process model or a process structure tree \cite{Goun} via intuitively creating nodes and building sequence flows from the extracted procedural knowledge \cite{process_model_generation}.

\subsection{Coarse-grained Learning of the Sentence-Level Knowledge}
The goal of this phase is to learn sentence-level knowledge in advance. First, it takes a sequence of embedded words $S^i=\langle W^i_1, W^i_2, \cdots \rangle$ as input. Then, the word vectors are bidirectionally encoded to capture inherent clues in a sentence. Meanwhile, a convolutional neural network (CNN) is used to capture local $n$gram features in a sentence. After concatenating $n$gram features and sentence embedding, two sentence-level tasks are trained jointly, using a weight-sharing multitask learning framework, to share learned high-level knowledge between two tasks and improve the generalization ability. 

\subsubsection{Embedding Layer} We use BERT \cite{BERT} or Word2Vec to obtain word vectors. BERT is an autoencoding language model, which has been the currently state-of-the-art pre-training approach. Given the input token sequence, a certain portion of tokens are replaced by a special symbol [MASK], and the model is trained to recover the original tokens from the corrupted version. Since density estimation is not part of the objective, BERT is allowed to utilize bidirectional contexts for reconstruction. This can close the aforementioned bidirectional information gap in autoencoding language modeling, leading to improved performance.

\subsubsection{Encoding Layer} As a special type of recurrent neural network (RNN), LSTM \cite{LSTM} is particularly suitable for modeling the sequential property of text data. At each step, LSTM combines the current input and knowledge from the previous steps to update the states of the hidden layer. To tackle the gradient vanishing problem of traditional RNNs, LSTM incorporates a gating mechanism to determine when and how the states of hidden layers can be updated. Each LSTM unit contains a memory cell and three gates (i.e., an input gate, a forget gate, and an output gate). The input and output gates control the input activations into the memory cell and the output flow of cell activations into the rest of the network, respectively. The memory cells in LSTM store the sequential states of the network, and each memory cell has a self-loop whose weight is controlled by the forget gate. Let us denote each sentence as $(S_i, L_i)$, where $S_i = [W_i^1,W_i^2,\cdots,W_i^n]$ as a sequence of word vectors representing the plain text and $L_i$ as its label; and $[d^1,d^2,\cdots,d^n]$ denotes the sequence of word vectors of $S_i$. At step $t$, LSTM computes unit states of the network as follows:
\begin{align}
i^{(t)}=&\sigma(U_id^t+W_ih^{(t-1)}+b_i)\\
f(t)=&\sigma(U_fd^t+W_fh^{(t-1)}+b_f)\\
o(t)=&\sigma(U_od^t+W_oh^{(t-1)}+b_o)\\
c(t)=&f_t\odot c^{(t-1)}+i^{(t)}\odot \tanh (U_cd^t+W_ch^{(t-1)}+b_c)\\
h^{(t)}=&o^{(t)}\odot\tanh (c^{(t)})
\end{align}
where $i^{(t)}$, $f^{(t)}$, $o^{(t)}$, $c^{(t)}$, and $h^{(t)}$ denote the state of the input gate, forget gate, output gate, memory cell, and hidden layer at step $t$. $W$, $U$, $b$ respectively denote the recurrent weights, input weights, and biases. $\odot$ is the element-wise product. We can extract the latent vector for each step $T$ from LSTM. In order to capture the information from the context both preceding and following a word, we use the bi-directional LSTM (Bi-LSTM) \cite{BiLSTM}. We concatenate the latent vectors from both directions to construct a bi-directional encoded vector $h_i$ for every single word vector $W_i^j$, which is:

\begin{align}
\overrightarrow{h_i}=&\overrightarrow{LSTM}(W_i^j), i \in [1,|S_i|]\\
\overleftarrow{h_i}=&\overleftarrow{LSTM}(W_i^j), i \in [1,|S_i|]\\
h_i=&[\overrightarrow{h_i},\overleftarrow{h_i}]
\end{align}

\subsubsection{Convolution Layer}
We also employ multiscale filters to capture local $n$gram information in a sentence. Let $x_1^n$ refer to the concatenation of vectors $x_1, x_2, \cdots, x_n$. The convolution layer involves a set of filters $w \in \mathbb{R}^{h \times k}$, which is solely applied to a window of $h$ to produce a new feature map $v=$:
\begin{equation}
\begin{bmatrix}
\begin{bmatrix}
\sigma (w_i x_1^h + b_1)\\
\sigma (w_i x_2^{h+1} + b_1)\\
\cdots \\
\sigma (w_i x_{n-h+1}^n + b_1)\\
\end{bmatrix}
,
\begin{bmatrix}
\sigma (w_i x_1^h + b_2)\\
\sigma (w_i x_2^{h+1} + b_2)\\
\cdots \\
\sigma (w_i x_{n-h+1}^n + b_2)\\
\end{bmatrix}
\cdots
\end{bmatrix}
\end{equation}where $\sigma(\cdot)$ is the sigmoid function and $b_i$ is a bias term. Meanwhile, we use the max-pooling operation: $\hat{v}=max(v_i)$ to extract the most important features within each feature map.

\subsubsection{Multi-Task Learning}
The goal of this module is to incorporate the multiple features for final sentence-level predictions. Since not all features contribute equally to the final task, we employ the gate attention mechanism to weight each concept information. For a feature representation $z$, we define the gate-attention as follows: $g = \sigma( W z + b )$, where $\sigma$ denotes the sigmoid activation function, which guarantees the values of $g$ in the range $[0,1]$. $W$ and $b$ are a weight and a bias term which need to be learned. Then, all convoluted representations are injected into the sentence representation by weighted fusing: $z_c = z_T \oplus ( g \otimes z )$, where $\otimes$ denotes element-wise multiplication, $\oplus$ denotes concatenation and $z_T$ is the BiLSTM representation. The effect of gate attention is similar to that of feature selection. It is a ``soft'' feature selection which assigns a larger weight to a vital feature, and a small weight to a trivial feature. Note that we additionally incorporate the hidden feature of ST1 into the input representations of ST2 via a self-attention gate to model the conditional relation between the two sentence-level tasks.

After concatenating all features, it is input into two fully connected multi-layer perceptron (MLP) networks to realize feature fusion:
\begin{equation}
o_i = softmax(W_2 \cdot (W_1 \cdot V + b_1) + b_2)
\end{equation}
where $W_1$, $W_2$, $b_1$ and $b_2$ are parameters of a network. To obtain the probability distribution on each type $t\in[1,T]$, the \textit{softmax} operation is computed by: $p_k=\frac{\exp (o_t)}{\sum_{i=1}^T \exp (o_i)}$, where $T$ is the class number of a classification task.

\subsubsection{Model Training (The Coarse-grained Learning Phase)}
We use cross-entropy loss function to train the coarse-grained learning phase, when given a set of training data $x_t,y_t,e_t$, where $x_t$ is the $t$-th training example to be predicted, $y_t$ is one-hot representation of the ground-truth type and $e_t$ is the model's output. The goal of training is to minimize the loss function:
\begin{equation}
\begin{aligned}
J(\theta_1,\theta_2) = &- \lambda _1 \big ( \sum_{i=1}^M \sum_{t_1=1}^{t_1=T_1} y_t^{t_1} \cdot \log (e_t^{t_1}) \big ) - \lambda _2 \big (\sum_{i=1}^M \sum_{t_2=T_1+1}^{t_2=T_1+T_2} y_t^{t_2} \cdot \log (e_t^{t_2}) \big ) \\
& \lambda _1 + \lambda _2= 1, \lambda _1, \lambda _2\ge 0
\end{aligned}
\end{equation}
where $M$ is the number of training samples; $T_{1/2}$ is the category number of each subtask; $\lambda_{1/2}$ is a linear balance parameter. The multi-task learning phase can be further decomposed into learning two single tasks successively.

\subsection{Fine-grained Learning of the Word-Level Knowledge}
Note that the features of a small-scale classification task may be higher quality than the features of a large-scale classification tasks since the small-scale classification task has relatively sufficient training data \cite{From_Small_scale}. Based on that, compared with sentence-level tasks, we regard the word-level tasks as relatively ``large-scale'' classification tasks, initializing parameters of the word-level task from the sentence-level tasks and later fine-tuning it. The transfer of parameters can provide a better starting point for the latter than training from scratch using randomly initialized parameters \cite{Towards_Unsupervised_Text}.

As the right part of Figure \ref{fig:Framework} shows, we extract the last-hidden features $z_s$ in ST1 as learned sentence-level knowledge and concatenate it with word-level embedding $z_w$ via the gate-attention mechanism. The fused representation $[z_s,g(z_w)\odot z_w]$ is fed into a MLP module to perform the word-level prediction. In this phase, we freeze ST2 (the light gray part) since ST2 and ST3 are independent.

\subsubsection{Model Training (The Fine-grained Learning Phase)}
Similarly, we use cross-entropy loss function to train the fine-grained learning phase, when given a set of training data $x_t,y_t,e_t$, where $x_t$ is the $t$-th training example to be predicted, $y_t$ is one-hot representation of the ground-truth type and $e_t$ is the model's output. The goal of training is to minimize the loss function:
\begin{equation}
\begin{aligned}
J(\theta_3) = &- \sum_{i=1}^M \sum_{t_3=1}^{t_3=T_3} y_t^{t_3} \cdot \log (e_t^{t_3})\\
\end{aligned}
\end{equation}
where $M$ is the number of training samples; $T_{3}$ is the category number of ST3.

\section{Experiments}
In this section, we first introduce the experimental setup (datasets, baselines and implementation details). The experimental results are then demonstrated to validate the effectiveness of our approach on different datasets.

\subsection{Datasets}
Since traditional methods mainly use off-the-shelf NLP tools to analyze sentences and extract corresponding information under the pre-defined features or patterns; thus, there was no directly available multi-grained corpus for the PME task. To this end, we constructed two multi-grained PME corpora for the task of extracting process models from texts:
\begin{enumerate}[$\bullet$]
\item \textbf{Cooking Recipes (COR).} We collected cooking recipes from the world’s largest food-focused social network\footnote{\url{https://www.recipe.com}}. This corpora has a large and diverse collection of more than 200 cooking recipes and covers every kind of meal type including appetizers, breakfast, desserts, drinks, etc.
\item \textbf{Maintenance Manuals (MAM).} We collected raw maintenance manuals from a wiki-based site\footnote{\url{https://www.ifixit.com}} that teaches people how to fix almost all devices. This corpora contains more than 160 maintenance descriptions and covers almost all devices including computer hardware, phones, cars, game consoles, etc.
\end{enumerate}

In the two raw corpora, we first split all documents into sentences and manually assigned labels to them. The sentence-level tags denote whether a sentence is describing cooking ingredients or maintenance tools, performing main actions, or executing conditions. Furthermore, we split all sentences into words and manually assign labels to them to denote the executor, action name and ingredients/tools of actions, i.e., the semantic roles. The statistics of those datasets is given in Table \ref{tab:statistics}.

\begin{table}[htbp]
\setlength{\abovecaptionskip}{0pt}
\setlength{\belowcaptionskip}{0pt}
\small
\centering
\caption{Statistics of the two multi-grained datasets.}
\label{tab:statistics}
\begin{tabular}{|c|c|c|}
\hline
& \textbf{COR} & \textbf{MAM} \\
\hline
Domain & Recipe & Maintenance \\
\hline
\# Labeled Sentences & 2,636 & 2,172 \\
\hline
\# Labeled Words & 14,260 & 20,612 \\
\hline
\# Sentence-Level Categories & 5 & 5 \\
\hline
\# Word-Level Categories & 4 & 4 \\
\hline
\end{tabular}
\end{table}

\subsection{Baselines}
We chose several representative baselines:
\begin{enumerate}[$\bullet$]
\item A pattern-matching based method (PBSW) \cite{extracting_and_enriching}, which uses NLP tagging tools to extract important linguistic patterns and adopts a set of heuristic anaphora resolution mechanisms to extract corresponding procedural knowledge.
\item A language-rule based method (ARWE) \cite{on_the_Use}, which introduces a lexical approach and two further approaches based on a set of association rules created during a statistical analysis of a corpus of workflows.
\item A traditional-learning based method (RPASVM) \cite{identifying_candidate_tasks}, which leverages SVM to automatically identify whether a task described in a textual description is manual, an interaction of a human with an information system or automated.
\end{enumerate}

\subsection{Implementation Details}\label{ssec:implementation}
We used BERT \cite{BERT} with text representations of size 100 (also can use Word2Vec \cite{Word2Vec} in our architecture). In training, we used a Rectified Linear Unit (ReLU) as an activation function \cite{imagenet_classification} and the Adam optimizer \cite{adam}. The network was run for 1,000 iterations with a mini-batch size of 32 at the coarse and the fine training. The learning rate is $10^{-4}$. We implement MGTC using Python 3.7.3\footnote{\url{https://www.python.org}} and Tensorflow 1.0.1\footnote{\url{https://www.tensorflow.org}}. All of our experiments were run on a single machine equipped with an Intel Core i7 processor, 32GB of RAM, and an NVIDIA GeForce-GTX-1080-Ti GPU. For comparison, all methods have been evaluated with the same training and test data. We divide datasets into train/test sets using an 8:2 ratio. The statistical significance between all baseline methods and MGTC is tested using a two-tailed paired t-test \cite{The_Hitchhikers_Guide}. Our code and data are available at \url{https://github.com/qianc62/MGTC}.

\subsection{Overall Performance}
We compare all baseline methods, MGTC and its variants in terms of \emph{classification accuracy}. Besides, we extract corresponding process models from the classification results and compare the \textit{behavior similarity} between extracted models and the gold models using existing behavior evaluation methods \cite{Efficient_Consistency_Measurement}. The overall results (accuracy of predicted labels) are summarized in Table \ref{tab:accuracy}. From the evaluation results, we can make several key observations: 

\begin{table*}[t]
    \scriptsize
    \centering
    \caption{Experimental results (accuracy; \%) of all baselines, MGTC and its variants. \emph{DOP}, \emph{PPP}, \emph{TVC}, \emph{OPM}, \emph{ARM} and \emph{FMS} denote mechanisms applied in reference papers. $\blacktriangle$ and $\triangle$ indicate the best and the second-best performing methods among all the baselines, respectively. The best performance among all methods is highlighted in boldface. The * denotes statistical significance ($p \le 0.05$) compared to MGTC.}
    \label{tab:accuracy}
    \begin{tabular}{|l||ll|ll|ll||ll|}
    \hline
    \multirow{2}*{\diagbox[width=15.6em,height=2.0em]{Method}{Task \& Dataset}} & \multicolumn{2}{c|}{\textbf{ST1}} & \multicolumn{2}{c|}{\textbf{ST2}} & \multicolumn{2}{c||}{\textbf{ST3}} & \multicolumn{2}{c|}{\textbf{PME}} \\
    \cline{2-9}
    & \textbf{ COR} & \textbf{MAM} & \textbf{COR} & \textbf{MAM} & \textbf{COR} & \textbf{MAM} & \textbf{COR} & \textbf{MAM} \\
    \hline
    \hline
    \textbf{PBSW}+\emph{DOP} & 47.63$_*$ & 47.58$_*$ & 43.66$_*$ & 39.49$_*$ & 34.38$_*$ & 35.60$_*$ & 30.35$_*$ & 30.27$_*$ \\
    \textbf{PBSW}+\emph{DOP}+\emph{PPP} & 51.52$_*$ & 52.55$_*$ & 45.40$_*$ & 45.67$_*$ & 48.58$_*$ & 47.36$_*$ & 38.04$_*$ & 37.17$_*$ \\
    \textbf{PBSW}+\emph{DOP}+\emph{PPP}+\emph{TVC} & 66.53$_*$ & 47.42$_*$ & 55.48$_*$ & 46.65$_*$ & 50.28$_*$ & 49.62$_*$ & 46.27$_*$ & 36.07$_*$ \\
    \textbf{ARWE}+\emph{OPM} & 70.66$_*$ & 60.46$_*$ & 62.43$_*$ & 62.45$_*$ & 58.46$_*$ & 61.62$_*$ & 52.46$_*$ & 51.09$_*$ \\
    \textbf{ARWE}+\emph{OPM}+\emph{ARM} & 76.39$_*$ & 69.51$_*$ & 56.52$_*$ & 72.53$_*^\triangle$ & 71.38$_*^\triangle$ & 77.62$_*^\triangle$ & 65.21$_*^\triangle$ & 59.50$_*$ \\
    \textbf{ARWE}+\emph{OPM}+\emph{ARM}+\emph{FMS} & 71.63$_*$ & 69.60$_*$ & 61.37$_*$ & 64.48$_*$ & 72.64$_*^\blacktriangle$ & 78.26$_*^\blacktriangle$ & 69.57$_*^\blacktriangle$ & 66.11$_*^\blacktriangle$ \\
    \textbf{RPASVM}+\emph{1}gram & 82.53$_*^\triangle$ & 81.65$_*^\triangle$ & 84.44$_*^\triangle$ & 71.55$_*$ & 63.39$_*$ & 60.39$_*$ & 57.76$_*$ & 60.23$_*$ \\
    \textbf{RPASVM}+\emph{1}gram+\emph{2}gram & 85.37$_*^\blacktriangle$ & 84.50$_*^\blacktriangle$ & 87.61$_*^\blacktriangle$ & 72.55$_*^\blacktriangle$ & 67.33$_*$ & 64.53$_*$ & 56.59$_*$ & 61.23$_*^\triangle$ \\
    \hline
    \hline
    \textbf{MGTC} & \textbf{93.34} & \textbf{91.74} & \textbf{91.53} & \textbf{86.49} & \textbf{82.39} & \textbf{80.44} & \textbf{77.40} & \textbf{75.77} \\
    \textbf{MGTC}$\diagdown$Gate Mechanism & 90.37$_*$ & 88.58$_*$ & 89.40$_*$ & 84.57$_*$ & 77.62$_*$ & 76.42$_*$ & 67.39$_*$ & 71.64$_*$ \\
    \textbf{MGTC}$\diagdown$Coarse-to-Fine & 91.55$_*$ & 89.45$_*$ & 88.31$_*$ & 84.53$_*$ & 79.44$_*$ & 76.22$_*$ & 74.46$_*$ & 72.29$_*$ \\
    \hline
    \end{tabular}
    \end{table*}

\begin{enumerate}[1)]
\item Our proposed model MGTC consistently outperforms all methods on all subtasks with statistical significance. On ST1 (single sentence classification), MGTC improves the accuracy by 7.97\% and 7.24\% on COR and MAM against the strongest competitor, respectively. On ST2 (sentence semantics recognition), MGTC improves the accuracy by 3.92\% and 13.94\% on COR and MAM, respectively. On ST3 (semantic role labeling), MGTC still outperforms all methods, improving accuracy by 9.75\% and 2.18\%, respectively. We believe that this is promising as word-level tasks face the problem of sparseness and ambiguity compared with sentence-level tasks, i.e., words have not relatively enough contextual information, which poses a great challenge for ST3. Moreover, in terms of behavior similarity between the extracted models and the gold models, employing the deep-learning-based framework improves behavior accuracy by 7.83\% and 12.66\% respectively, which further verifies that MGTC can extract procedural knowledge without employing manual features and complex procedural knowledge.
\item From ablation studies, we can see that the performance improves after employing the gate-attention mechanism and coarse-to-fine learning. For example, employing coarse-to-fine learning improves the behavior similarity by 2.94\% and 3.48\% on COR and MAM. That is to say, both the gate-attention mechanism and coarse-to-fine learning can improve accuracy, which shows the effectiveness of the two mechanisms. This is consistent with our intuition that training coarse tasks in advance before fine-grained tasks can learn to better learning procedural knowledge.
\item The experimental results also demonstrate the disadvantages of the manual-feature-based methods. First, since traditional methods employ diverse manual mechanisms or features, they suffer from the problems of poor quality and lack of adaptability. In contrast, deep-learning-based methods learn hidden contextual features automatically and always maintain their superior performance. Second, by considering the long-term language changes, the framework with mixture of multiscale filters can understand texts' semantics better and extract more accurate information. Third, over all subtasks and datasets, we can see that MGTC can maintain more stable results, which suggests that MGTC is more robust on different subtasks and datasets.
\end{enumerate}

\subsection{Further Investigation}
\begin{figure*}[h]
\centering
\subfigure[The effect of language models.]{
\label{sfig:ablation_model}
\includegraphics[width=0.3\columnwidth]{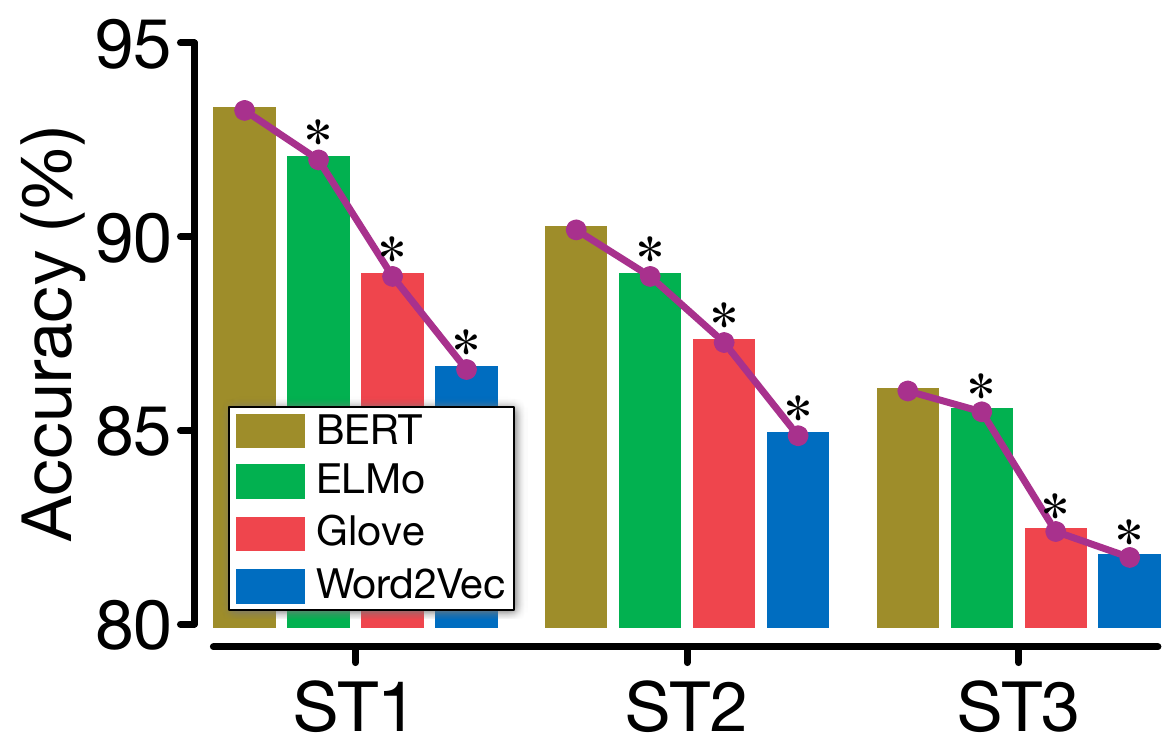}
}
\subfigure[The effect of $n$gram filters]{
\label{sfig:ablation_gram}
\includegraphics[width=0.3\columnwidth]{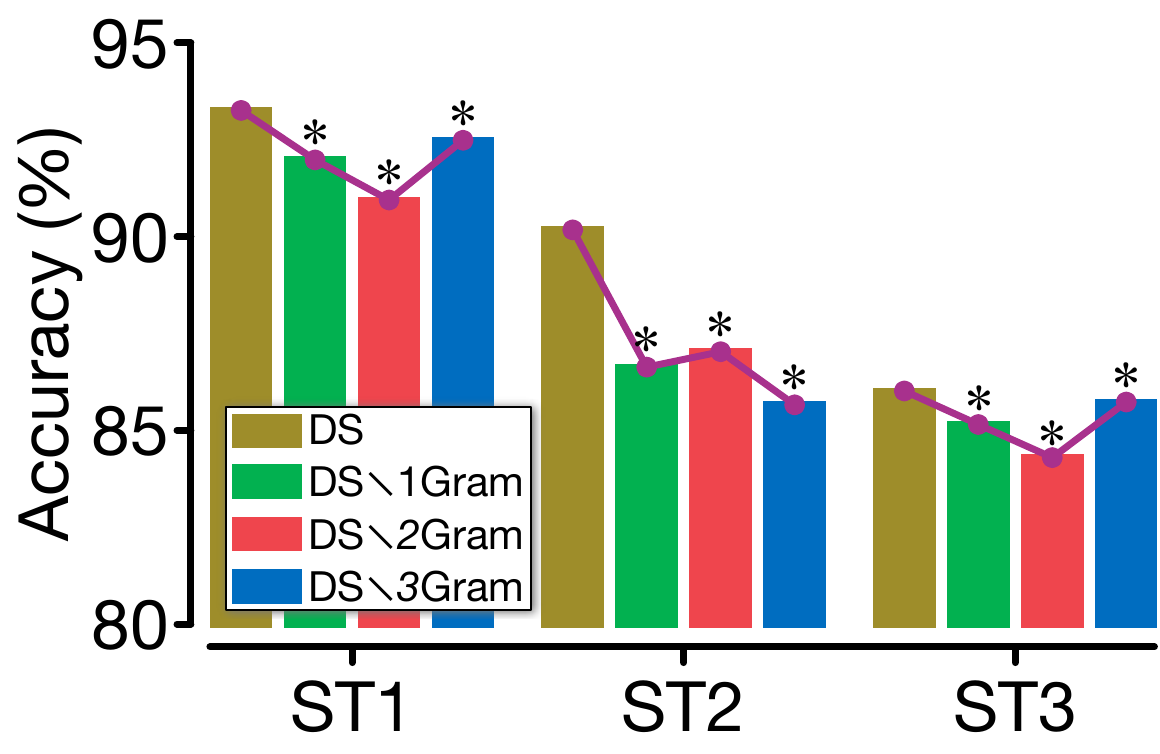}
}
\subfigure[The effect of \#hidden layers]{
\label{sfig:ablation_layer}
\includegraphics[width=0.3\columnwidth]{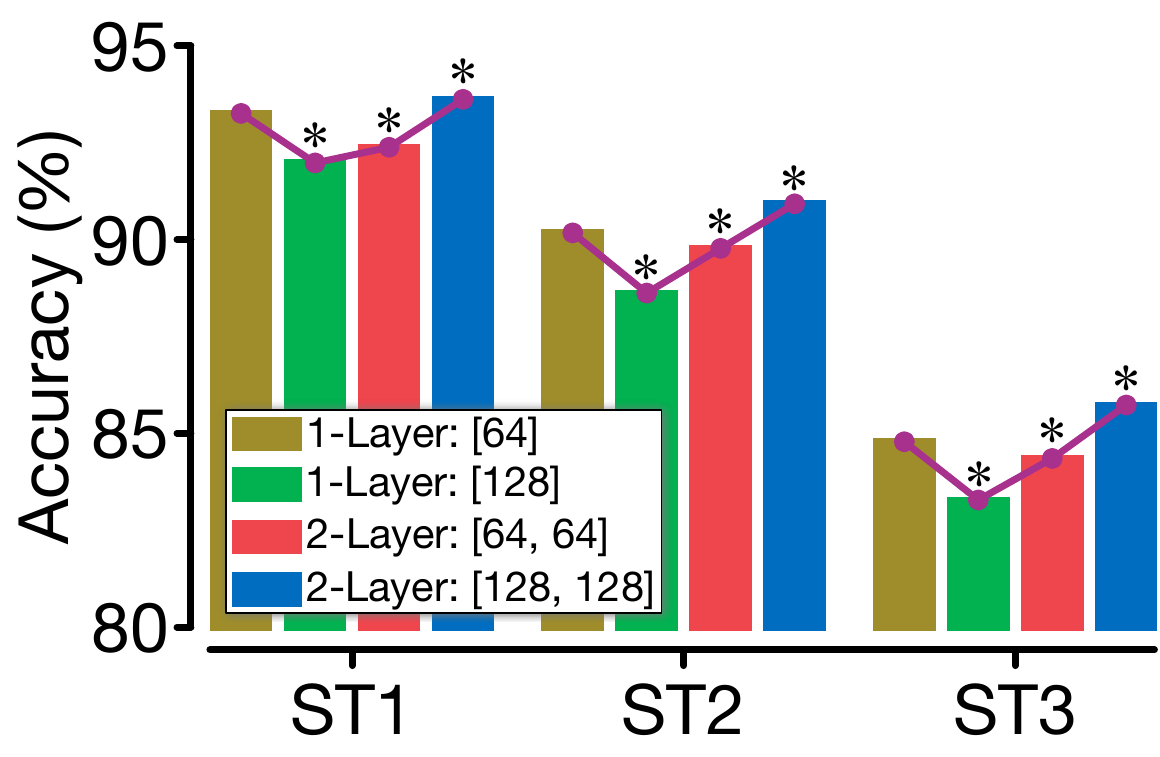}
}
\caption{Ablation analysis of different components (language models, $n$gram filters, the number of hidden layers) on three subtasks. DS denotes the default settings. $\diagdown$ is the removing operation. * denotes statistical significance ($p \le 0.05$) compared to MGTC.}
\label{fig:ablation}
\end{figure*}

To further investigate the independent effect of the key parameters or components in our framework, we compare our method with those replacing with other standard components, using the default settings described in Section \ref{ssec:implementation}. The results are shown in Figure \ref{fig:ablation}.

\textbf{Effect of Language Models.} Figure \ref{sfig:ablation_model} shows the performance of those variants employing different language models (BERT, ELMo, GloVe and Word2Vec). It can be observed that, when the default BERT is replaced with ELMo, GloVe or Word2Vec, the performance consistently decreases. Particularly, on ST3, using Word2Vec decreases the accuracy about 4.58 points, which shows that pre-training on large-scale corpora can better capture linguistic information.

\textbf{Effect of Multi-scale $\bm{n}$gram.} Figure \ref{sfig:ablation_gram} shows the performance of the variants after removing $n$gram ($n$=1,2,3) filters from MGTC. It can be observed that, when $n$gram filters are removed from the network, it decreases performance of on average by 3.18\%/4.02\%/1.06\% on the three tasks respectively. This demonstrates that $n$gram filters make a contribution to the promising scores. In particular, removing \emph{3}gram filters obviously influences the performance of ST2, suggesting that \emph{3}gram features play an important role in ST2.

\textbf{Effect of The Number of Hidden Layers.} From Figure \ref{sfig:ablation_layer}, we can observe that the performance of two-layer MLP indicates that imposing some more non-linearity is useful, and that the one-layer networks seem slightly weak for certain subtasks. However, further increasing the number of layers may degrade the performance because the model becomes too complicated to train. Using a two-layer architecture seems like a good trade-off between speed and accuracy.

\begin{figure*}[h]
\centering
\subfigure[Time consumption.]{
\label{sfig:Time}
\includegraphics[width=0.40\columnwidth]{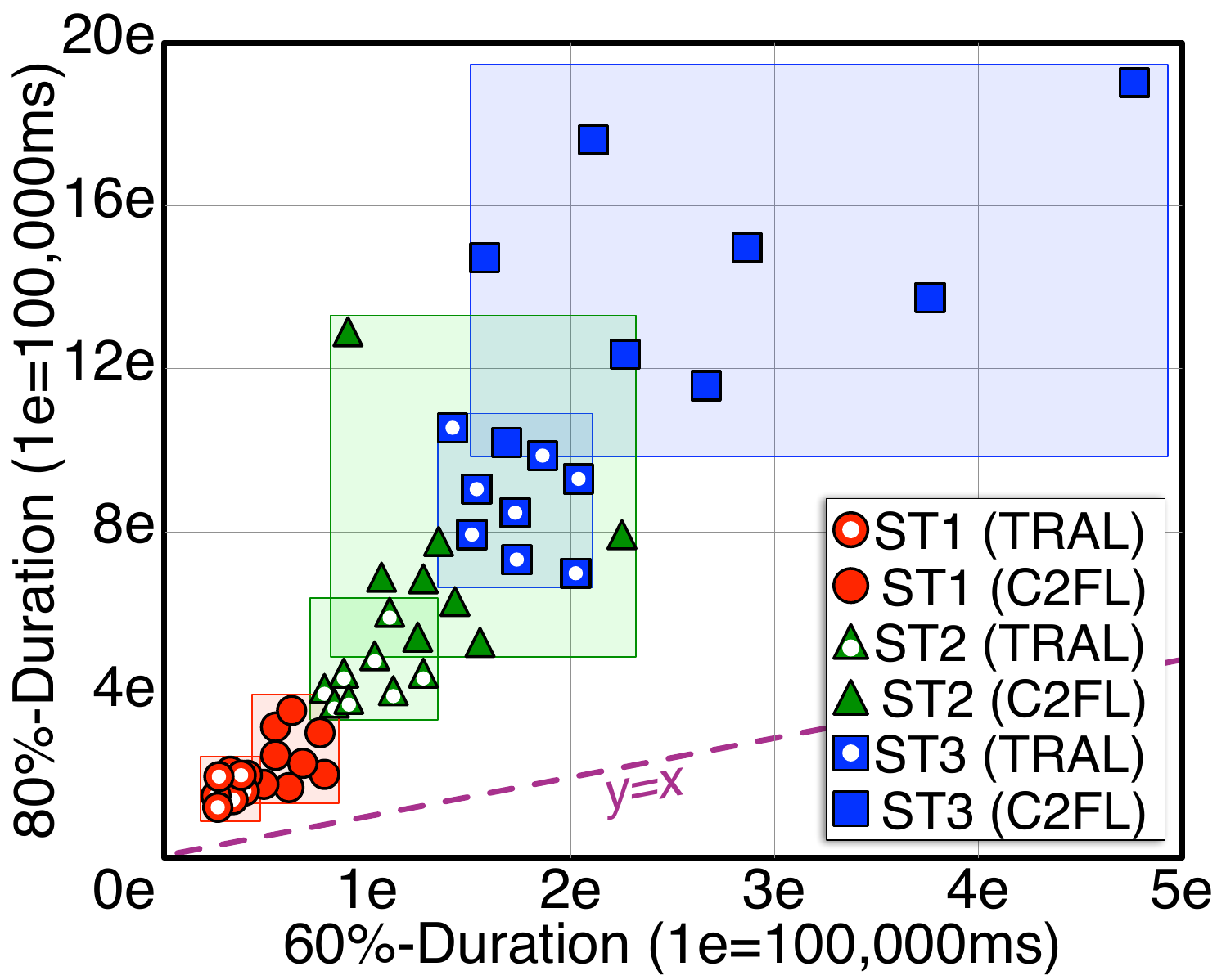}
}
\hspace{0.2cm}
\subfigure[N-fold (N=2,3,$\cdots$,20) cross validation results.]{
\label{sfig:Nfold}
\includegraphics[width=0.40\columnwidth]{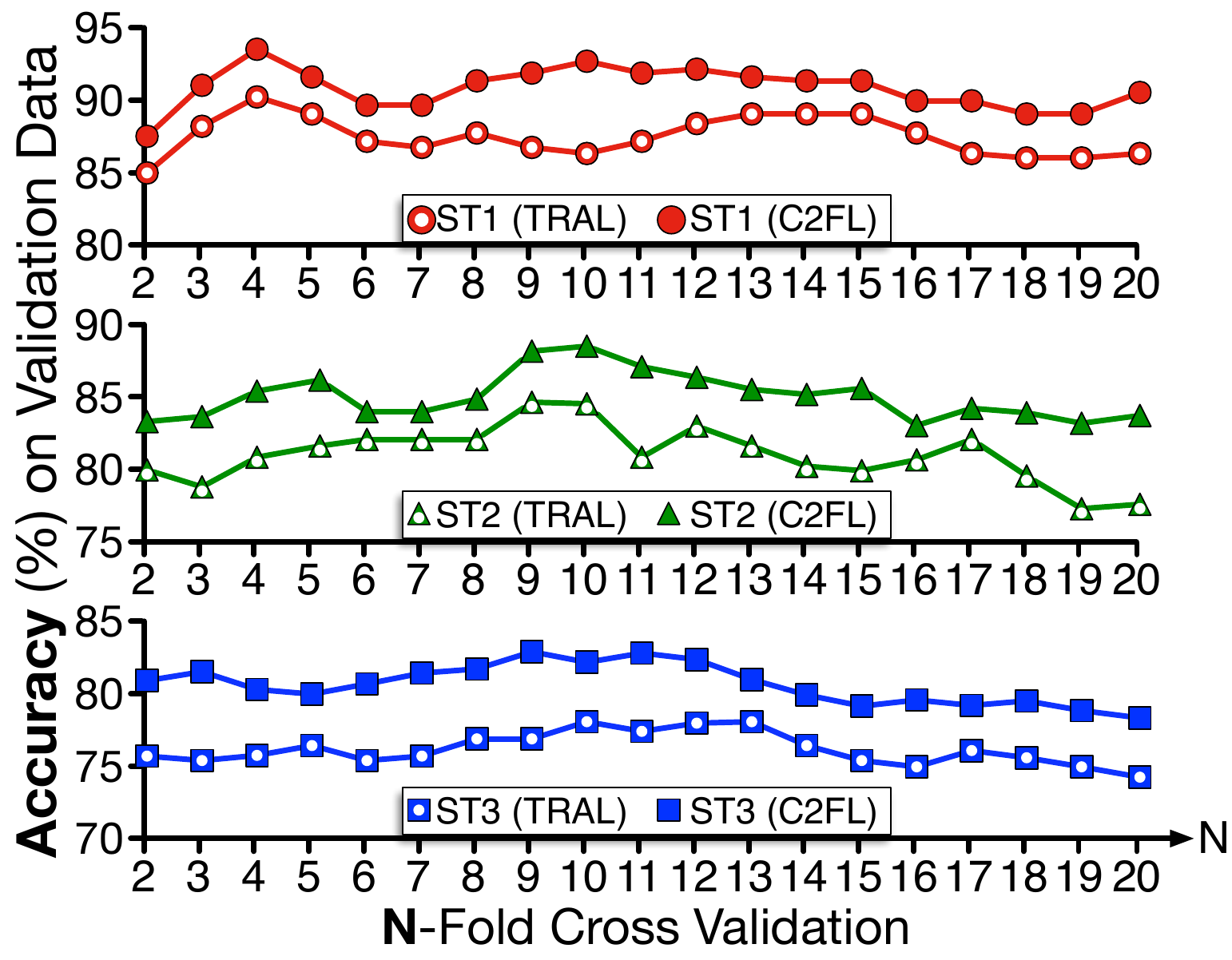}
}
\caption{Further investigation of the traditional learning (TRAL) and our proposed coarse-to-fine learning (C2FL).}
\end{figure*}

We also conducted more experiments to further investigate the performance of traditional learning (TRAL) and our proposed coarse-to-fine learning method (C2FL). From Figure \ref{sfig:Time} we can observe that ST1 first converges after a $1.8e$ average duration ($1e=10^5ms$), followed by ST2 ($4.1e$) and ST3 ($8.2e$). Moreover, comparing the results of TRAL and C2FL, one can see that C2FL consistently magnifies and scatters the time consumption on three subtasks. The main reason for this is that coarse-to-fine learning architecture would leverage learned features from sentence-level tasks to word-level tasks in a gradual manner; thus, word-level tasks (ST3) would consume more time than sentence-level tasks (ST1 and ST2). From Figure \ref{sfig:Nfold}, the N-fold cross-validation results show that the generalization performance of C2FL is remarkably higher than TRAL's on all subtasks. This further indicates that employing coarse-to-fine learning is more useful in extracting information with different granularities. Therefore, we can conclude that, although at the expense of longer running time, coarse-to-fine learning tends to converge to a higher performance. 

\section{Discussion}
Note that many publicly-available model-text pairs (e.g., \cite{process_model_generation}) are not considered as our datasets, because our supervised method inevitably needs \textbf{much-larger} scale of \textbf{labeled} (both sentence- and word-level) training data than traditional rule-based ones. Both requirements (data scale and label) are not fully satisfied on previous datasets which are almost uniformly small or unlabeled.

Moreover, the datasets used in this paper consist of relatively general process texts. Thus, a classic pattern-matching based method \cite{process_model_generation} is not considered in this version because it mainly focuses on extracting Business Process Model and Notation (BPMN) that contains many \textbf{specific} elements, such as swimlanes and artifacts, whose attributes currently have no corresponding specifically-defined labels to perform procedural knowledge extraction via our ``relatively general'' model. Adding such features is a possible extension of our work.

It is also worth mentioning that although we aim to extract from general process texts, our model might suffer from low external validity. One main reason is that both recipes and manuals are rather sequential by nature, which means that we can expect a rather small number of concurrent streams of actions and complex choices. However, note that deep-learning-based methods, to some extent, tend to be more generalized than pattern-based ones, due to their ability for ``adaptive knowledge learning'' (rather than those that work with a fixed human-defined knowledge). This limitation further motivates us to design more general and robust models in future work.

\section{Conclusion}
In this paper, we formalize the PME task into the multi-grained text classification problem and propose a hierarchical multi-grained network to model and extract multi-grained information without manually defined procedural features. Under this structure, we accordingly propose the coarse-to-fine learning mechanism, training multi-grained tasks in coarse-to-fine grained order, to apply the high-level knowledge for the low-level tasks. The results demonstrate that our approach outperforms the state-of-the-art methods with statistical significance and the further investigations demonstrate its effectiveness. Therefore, we draw two main conclusions as follows: 1) The deep-learning-based process extractor can effectively capture procedural knowledge without the need for defining domain-specific procedural features; and 2) Our proposed hierarchical network and the coarse-to-fine learning mechanism can better learn the word-level clues based on pre-learned sentence-level knowledge.

\section*{Acknowledgement}
The work was supported by the National Key Research and Development Program of China (No. 2019YFB1704003), the National Nature Science Foundation of China (No. 71690231) and Tsinghua BNRist. Lijie Wen is the corresponding author.

\bibliographystyle{splncs04}
\bibliography{_manuscript}

\end{document}